\documentclass{llncs}
\input{psfig.sty}
\begin{document}
\pagestyle{empty}  
\title{Affine Invariant, Model-Based Object Recognition Using Robust Metrics and Bayesian Statistics}
\titlerunning{Object Recognition}  
%
\author{Vasileios Zografos \and Bernard F. Buxton}
\authorrunning{Vasileios Zografos and Bernard F. Buxton}   
\tocauthor{Vasileios Zografos (University College London),
Bernard F. Buxton (University College London)}
\institute{Department of Computer Science, \\ University College London, Gower Street,\\ 
London WC1E 6BT, UK\\
\email{\{v.zografos, b.buxton\}@cs.ucl.ac.uk}
}
\maketitle              
\begin{abstract}
We revisit the problem of model-based object recognition for intensity images and attempt to address some of the shortcomings of existing Bayesian methods, such as unsuitable priors and the treatment of residuals with a non-robust error norm. We do so by using a reformulation of the Huber metric and carefully chosen prior distributions. Our proposed method is invariant to 2-dimensional affine transformations and, because it is relatively easy to train and use, it is suited for general object matching problems.
\end{abstract}
\section{Introduction}
In this paper we will examine the view-oriented case for model-based object recognition, in which 2-dimensional representations of 3-dimensional objects are used, called aspects or characteristic views \cite{KoeDoo1979}. Such methods have recently become quite popular because of their applicability in many areas and their ease of implementation, since they avoid storing and reconstructing a full 3d model. In addition, there is evidence to suggest that view-oriented representations are used by the human visual system for object recognition \cite {TarWil1998}.
The view-oriented object recognition problem for a single view can be formulated as follows: 
\begin{definition}
Suppose that we have a prototype template function $F_0$, an image function $I$ and a transformation $T$ that transforms the template as $F=TF_0$. The goal of object recognition is to minimise the expression:
\begin{equation}
\min_\xi S=\int_{R(\xi)}g(I(x),F(x))d^2x\;  ,
\label{eq:Object_recognition_equation}
\end{equation}
with respect to the transformation parameters $\xi$, where $g(.,.)$ is an error metric and $R$ the parameter space. If the minimum is less than or equal to some threshold $\tau$, then we have a match.
\end{definition}
The main problem that arises from this formulation is the determination of the parameters $\xi$ that minimise the above expression. Solving for $\xi$ depends on the transformation $T$. For complicated transformations $T$, the optimisation is a nonlinear process and the minimum is found using an iterative algorithm.
\subsection{Our Approach and Related Work}

We have based our approach on previous popular work by \cite{GreCho1991} and \cite{JaiZho1996}. First Grenander et al. proposed a general deformable template model, by representing deformations of the template as probabilistic transformations, for Bayesian inference on contour shape. Jain et al. used this approach together with a snake-like potential function to influence the template toward edge positions in the image. A similar scheme has been used by Cootes et al. \cite{CooEdw2001}, where the template is represented by the mean shape of a training set and a linear combination of the most important eigenmodes of the variation from the mean. The Bayesian object localisation method introduced by Sullivan et al. \cite{SulBla2001} is another interesting approach. Distributions of the template over the foreground and background are learned from training images, and used as the likelihood in a Bayesian inference scheme.

In our approach we use intensity information, without the need to extract features from the image. Also, we use novel distributions for the prior and do not assume that all transformations are equally possible. This disallows trivial solutions of the transformation parameters. Finally, our likelihood function is based on a more robust error metric that currently tends to one distribution when the template overlaps with an object (foreground) and to another when the template is on the background. A Bayesian formulation, that combines this prior knowledge together with information from the input image, the likelihood, is used in order to find a match between the image and the template. This combination is realised in the posterior probability, a maximum of which may indicate a possible match.
\section{Deformation Model}
The deformation model we propose consists of a prototype model template of the representative shape of an object, a selection of parametric transformations that act on the template, and a set of constraints that bias the choices of possible deformation parameters.
\subsection{Prototype Template Representation}
The prototype template consists of the pixels (grey levels) within a (for convenience) rectangular boundary, chosen as a representative example of an object or object class. The prototype is based on our prior knowledge about the objects of interest, and is usually obtained from training samples. Such training could be based on Principal Components Analysis (PCA), shape alignment, or the prototype could simply be the mean shape of the class. Unlike other methods, our model is not parameterised, but instead the transformation is. 
The model we are using contains grey level and boundary information in the form of a bitmap, and thus is appropriate for general object recognition tasks, since in order to apply the same method to a different class of objects we only need to generate a new prototype image of this class.
\subsection{Parametric Transformations}
Although the prototype template represents the most likely a-priori instance of the object, we still need to deform it to match the image. The parametric transformations consist of a global affine transformation $A$, and a local deformation $D$. 
It is necessary to compose $A$ as a product of individual meaningful transformations (primitive matrices). Such a composition is not unique, but by adopting a canonical order for the transformations, we could say, for example: $A=SRU_x+d$ where $S$ is an anisotropic scale matrix, $R$ is a rotation matrix, $U_x$ an angular shear matrix on the x-axis, and $d=(d_x,d_y)$ a translation vector. 

The local deformation $D$ is a 2d continuous mapping $(x,y)\rightarrow(x,y)+[D_{x}(x,y),D_{y}(x,y)]$, defined as a simple sinusoidal function:
\begin{equation}
D_{\psi}(x,y)=\left[D_{x}(x,y),D_{y}(x,y)\right]=\left[\alpha\cos(2\pi k_{x}\Delta),\beta\cos(2\pi k_{y}\Delta)\right]\;  ,\label{eq:LocalDeformation_NEW}
\end{equation}
where $\psi=(\alpha,\beta,k_{x},k_{y},x_{0},y_{0})$
are the deformation parameters, with $\alpha,\beta$ being the wave
amplitudes, $k_{x},k_{y}$ the wavenumbers, and $\Delta=\sqrt{(x-x_{0})^{2}+(y-y_{0})^{2}}$
is the Euclidean distance from the centre point $(x_{0},y_{0})$. 
We thus suppose that we have a prototype template function $F_{0}(x,y)$
and a transformation $T$ that transforms the template as follows: 
\begin{equation}
TF_{0}(x,y)=F_{0}(SRU_{x}(x,y)+D_{\psi}(x,y)+(d_{x},d_{y}))\;  .\label{eq:TotalTransformation}
\end{equation}
This is the parametric transformation that will deform the template
to match the image. This transformation is realised by shearing the
template by angle $\varphi$, then rotating by an angle $\vartheta$,
scaling the result by $s_{x}$, $s_{y}$ along directions $x$ and
$y$ respectively, locally deforming the resulting template by $\psi$
and finally a translation by $d$.
\subsection{Probabilistic Constraints}
Since not all choices of transformation parameters will produce a template that resembles the object(s) in the image, it is necessary to restrict their variability. We do so by imposing a probability density function (p.d.f) on the transformation $T$.

Consider the local deformation $D_{\psi}(x,y)$ first. We have chosen uniform distributions for the wave centre parameters $x_0$,$y_0$, since any centre point has an equal probability of producing a valid sinusoid. We further assume that the two sinusoids in (\ref{eq:LocalDeformation_NEW}) have amplitudes $\alpha$ and $\beta$ that are independently and identically normally distributed with zero mean and variance $\sigma_{\alpha\beta}^{2}$. For the wavenumbers $k_{x}$ and $k_{y}$, we also assume zero mean, independent and identical normal distributions with variance $w^2$. This results in a prior distribution for the shape parameters $\psi$:
\begin{equation}
Pr(\psi)=\frac{1}{4\pi\sigma_{\alpha\beta}^{2}w^2}\exp\left\{ -\frac{\alpha^{2}+\beta^{2}}{2\sigma_{\alpha\beta}^{2}}
-\frac{k_x^{2}+k_y^{2}}{2w^{2}}
\right\}\;  , \label{eq:DeformationD_pdf}
\end{equation}
that favours small deformations of the object in preference to large ones.

For the rotation and translation, we can assume that all rotations
and translations are equally possible and thus we can consider their
parameters $\vartheta,d$ as being uniformly distributed. However,
the scale and shear transformations require a different approach,
and special care is required for choosing their p.d.f.s. The reason
for this comes from the behaviour of the error function (\ref{eq:Object_recognition_equation}),
for certain values or ranges of values of the parameters $s=(s_{x},s_{y})$
and $\varphi$. More specifically, if one or both of the scale parameters are very
small, $F(x,y)$ will collapse into a single point or a line respectively. This
of course is not going to be a valid representation for the template
but the error function will undoubtedly have a minimum for these values
of the scale parameters. Such trivial solutions should not be allowed. Similar behaviour occurs with the shear angle $\varphi$, which for $\varphi=\pm\frac{\pi}{2}$, will collapse the object into a line.

To avoid these problems, we need to forbid such values for the scale and shear
parameters. To do so, we define a prior for these parameters that will bias them away from such values. A good choice for the scale parameters $s_{x}$ and $s_{y}$ is the inverse Gaussian (Wald) distribution \cite{EvaHas2000}, which, if we assume that
$s_{x}$ and $s_{y}$ are independent, that their mean scale $\bar{s}$
is $1$, and that their scale parameter is $\sigma_{s}$, leads to:
\begin{equation}
Pr(s)=\frac{\sigma_{s}\exp\left[-\frac{\sigma_{s}}{2}\left(-4+\frac{1}{s_{x}}+s_{x}+\frac{1}{s_{y}}+s_{y}\right)\right]}{2\pi\sqrt{s_{x}^{3}s_{y}^{3}}}\;  .\label{eq:Scale_pdf}
\end{equation}
The Wald distribution is ideal because it assigns very low probability to quantiles close to zero, while it allows us to determine the probability of large values of the scale parameter $s$ by adjusting the tail of the p.d.f.. 
For the shear angle, we would like to introduce a bias in favour of small deformations, and to rule out the values $\varphi=\pm\frac{\pi}{2}$. Furthermore, when the mean shear angle is zero, the distribution must be symmetric. On the other hand, if the mean angle is close to $-\frac{\pi}{2}$
then the distribution for negative values must fall sharply, whilst the
distribution for high values must exhibit similar behaviour when
the mean angle is close to (but not quite) $\frac{\pi}{2}$. We have therefore chosen a mixture model of two Gumbel distributions \cite{EvaHas2000}, with:
\begin{equation}
Pr(\varphi)=\frac{(1-A)e^{-\frac{\varphi-\overline{\varphi}}{b}-e^{-\frac{\varphi-\overline{\varphi}}{b}}}+Ae^{\frac{\varphi-\overline{\varphi}}{b}-e^{\frac{\varphi-\overline{\varphi}}{b}}}}{b}\;  ,\label{eq:Shear_pdf}
\end{equation}
where $b$ is the shape parameter and $A=\frac{\overline{\varphi}+\frac{\pi}{2}}{\pi}$.
Since the individual transformation parameters were assumed
independent, the total prior p.d.f. $Pr(\xi)$, is the product of the individual p.d.f.s (\ref{eq:DeformationD_pdf}),(\ref{eq:Scale_pdf}) and (\ref{eq:Shear_pdf}).
\section{Objective Function}
Two commonly used metrics in template matching applications are the $L_2$ metric and the $L_1$ metric which are valid from a maximum likelihood perspective, if the error residuals are normally distributed or exponentially distributed respectively. However, \cite {TiaYu2004} have shown that additive noise in real images is generally not normally distributed, and the majority of the variation comes from illumination changes and in-class object variation. In addition, \cite {SulBla2001} have shown that when using an error metric (such as the $L_2$) and considering only the portion of the image under the template, then the observations $I$ are a function of the hypothesis $\xi$. That is not valid in a Bayesian framework, since $I$ should be considered as fixed. In \cite {SulBla2001} a learning process is therefore used to model the different foreground and background distributions. Here, we use a simple parametric distribution to interpolate between the foreground and background behaviour. Since, in general we know little about the latter it should be based on a robust statistic. The $L_1$ metric, although robust, is singular when the residual goes to zero, and makes the optimisation process difficult. 
For this reason, we have chosen as a metric, a reformulation of the Huber norm \cite {Hub1973}. This {\it smooth Huber norm}, is $C^2$ continuous and defined as:
\begin{equation}
g_{\tau}(x)=\sqrt{1+\frac{x^{2}}{\tau^{2}}}-1\;  ,\label{eq:SmoothHuberNorm}
\end{equation}
where $\tau$ is the threshold between the L$_{\textrm{1}}$ and L$_{\textrm{2}}$
norms. The smooth Huber norm treats residuals close to zero (template over the foreground) with the $L_2$ norm and large residuals (template over the background) with the $L_1$ norm. By using equations (\ref{eq:Object_recognition_equation}), (\ref{eq:TotalTransformation}) and (\ref{eq:SmoothHuberNorm}) we obtain the combined objective function $S$ which needs to be minimized:
\begin{equation}
\min_\xi S(u,v)=\int_{{R(\xi)}}\left\{\sqrt{1+\frac{\left[I(u+x,v+y)-TF_{0}(x,y)\right]^{2}}{\tau^{2}}}-1\right\}\:\: dxdy\;  .\label{eq:Total_Objective_Function}\end{equation}
If we reformulate (\ref{eq:Total_Objective_Function}) as a p.d.f we see that the likelihood of observing the input image given the deformations on the prototype template is:
\begin{equation}
Pr(I|\xi)=C_{1}\exp\left\{ -S(u,v)\right\}\;  , \label{eq:Likelihood}\end{equation}where $C_{1}$ is a normalising constant, equal to $1/2(e K_{1}(1)\tau)$ where $e$ is the exponential and $K_{1}$ is a modified Bessel function. 

Finally, we may use the fact that $Pr(\xi|I)\propto Pr(I|\xi)Pr(\xi)$ and combine equations (\ref{eq:DeformationD_pdf}), (\ref{eq:Scale_pdf}), (\ref{eq:Shear_pdf}) and (\ref{eq:Likelihood}) to obtain the posterior p.d.f. of the parameters given an image $I$. The parameters may therefore be obtained by minimising the corresponding negative log-likelihood which for example, if the mean shear angle $\overline\varphi$ in (\ref{eq:Shear_pdf}) is zero, is given by:\[
\min_{\xi}\left\{-\log Pr(\xi|I)\right\}=
\log(\sqrt{s_{x}^{3}s_{y}^{3}})-\log\left(e^{-\frac{\varphi}{b}-e^{-\frac{\varphi}{b}}}+e^{\frac{\varphi}{b}-e^{\frac{\varphi}{b}}}\right)+\frac{k_{x}^{2}+k_{y}^{2}}{w^{2}}
\]
\begin{equation}
+\sigma_{s}\left(\frac{1}{s_{x}}+s_{x}+\frac{1}{s_{y}}+s_{y}-4\right)+\frac{\alpha^{2}+\beta^{2}}{\sigma_{\alpha\beta}^{2}}+S(u,v)\;  ,\label{eq:FINAL_OBJECTIVE_FUNCTION}\end{equation}
where $\xi=(s_{x},s_{y},\varphi,k_{x},k_{y},\alpha,\beta,x_{0},y_{0},d_{x},d_{y},\vartheta)$ are the transformation parameters. Note that the distribution shape parameters $b$, $w$, $\sigma_{s}$, $\sigma_{\alpha\beta}$ and the threshold $\tau$ are treated as constant.
\section{Experimental Results}
\begin{figure}  
\centerline{\begin{tabular}{ll}
  \psfig{figure=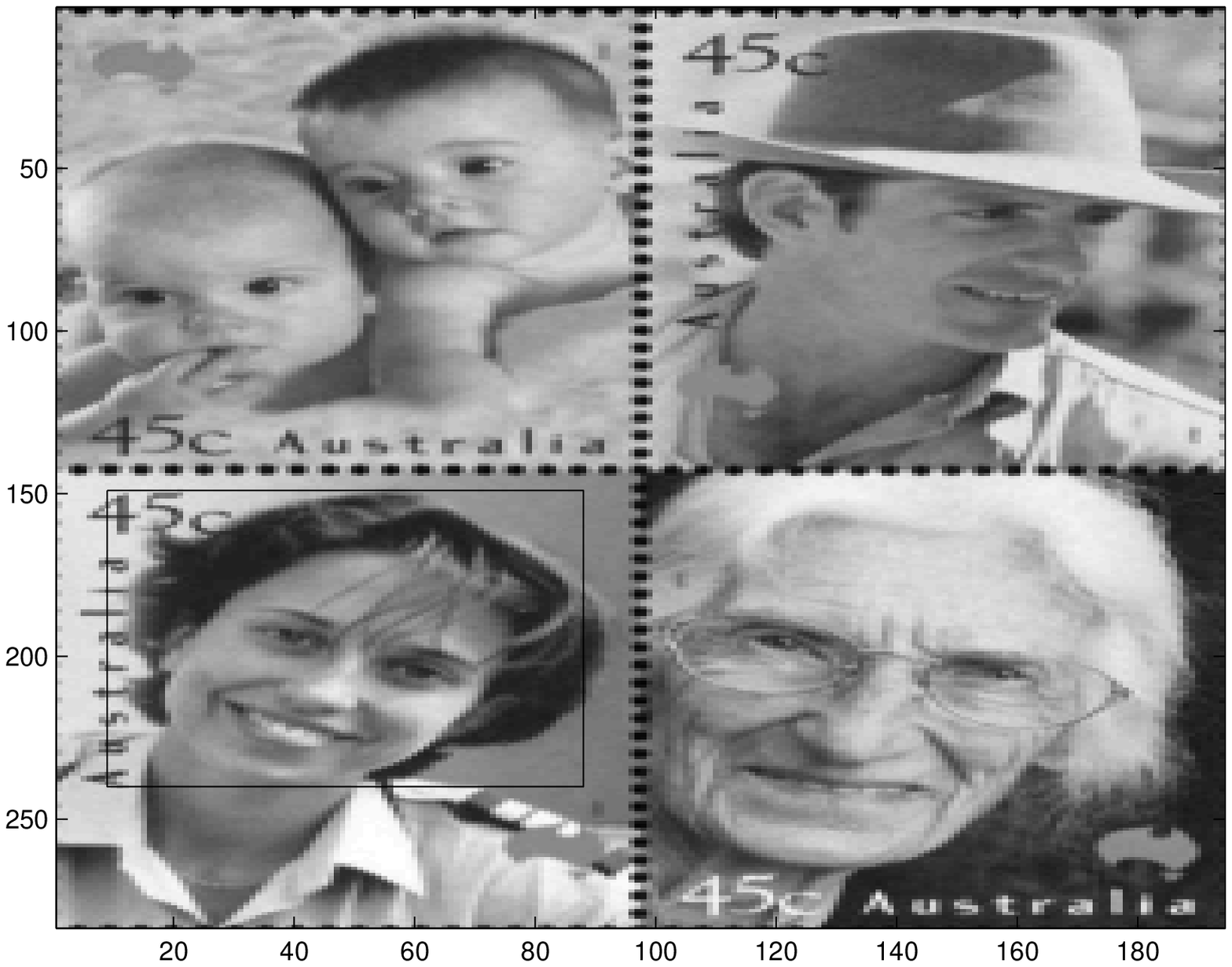,height=4cm}
  \hspace{10pt}
  &
  \psfig{figure=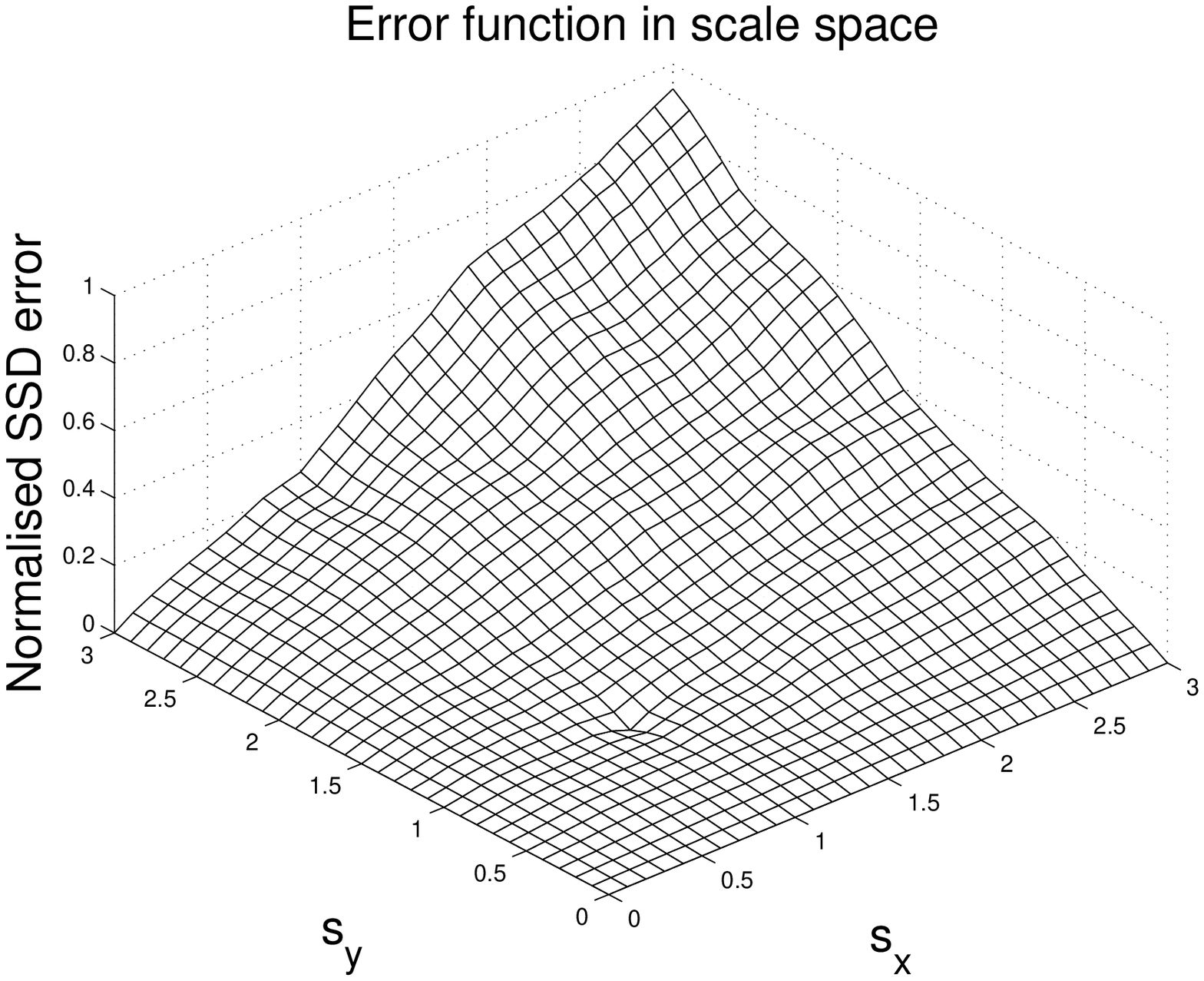,height=4cm}
\end{tabular}}
\vspace{5pt}
\centerline{\begin{tabular}{ll}
  \psfig{figure=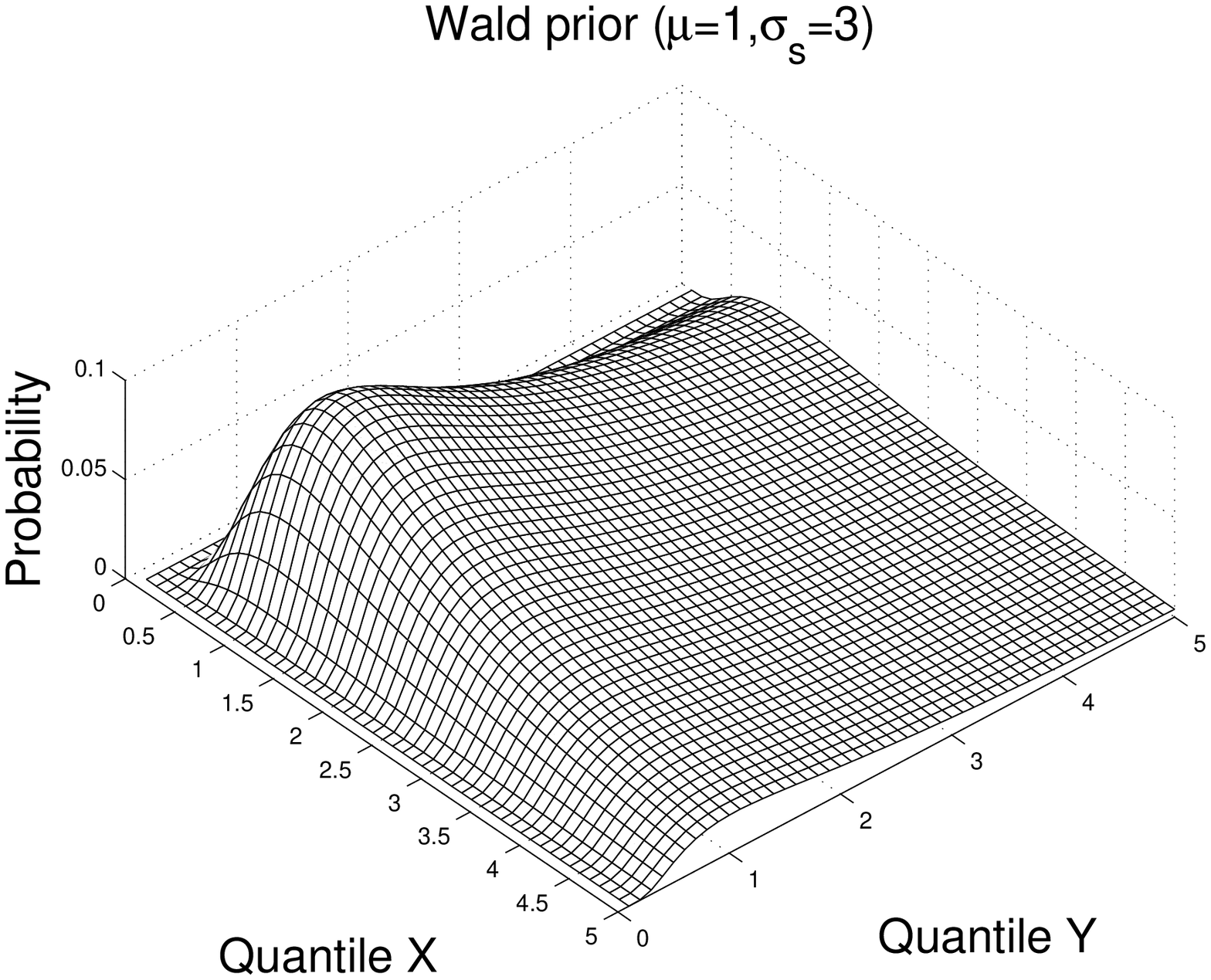,height=4cm}
  \hspace{10pt}
  &
  \psfig{figure=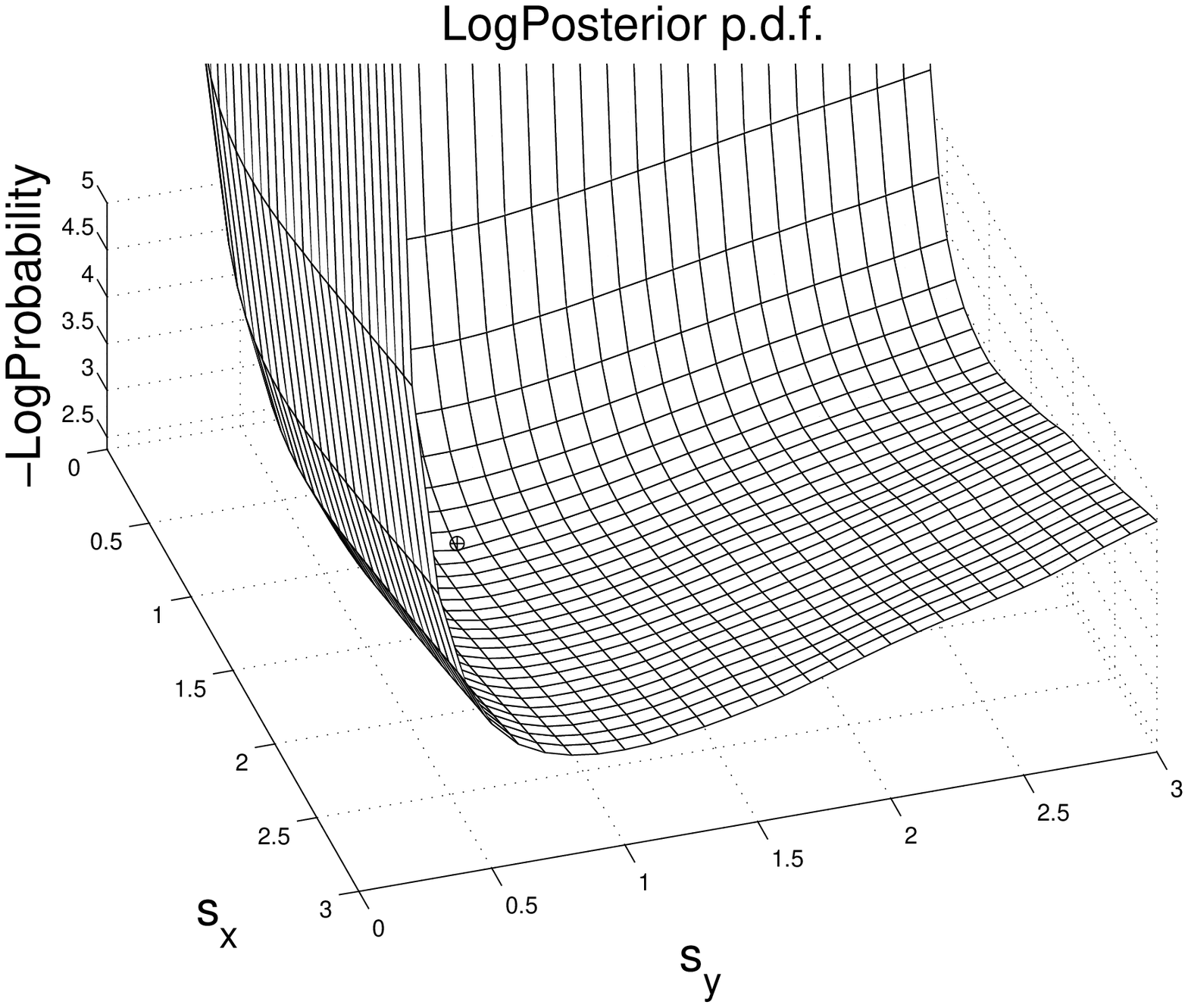,height=4cm}
\end{tabular}
}\caption{The test image and template ({\it top left}) and the error function for the scale parameters ({\it top right}). The desired solution is at $s_x=s_y=1$. The chosen Wald prior is illustrated ({\it bottom left}) and the resulting negative log-posterior probability ({\it bottom right}). The desired solution remains at the same position but without the trivial solutions}
\label{fig:Results1}
\end{figure}

\begin{figure}
\centerline{\begin{tabular}{ll}
  \psfig{figure=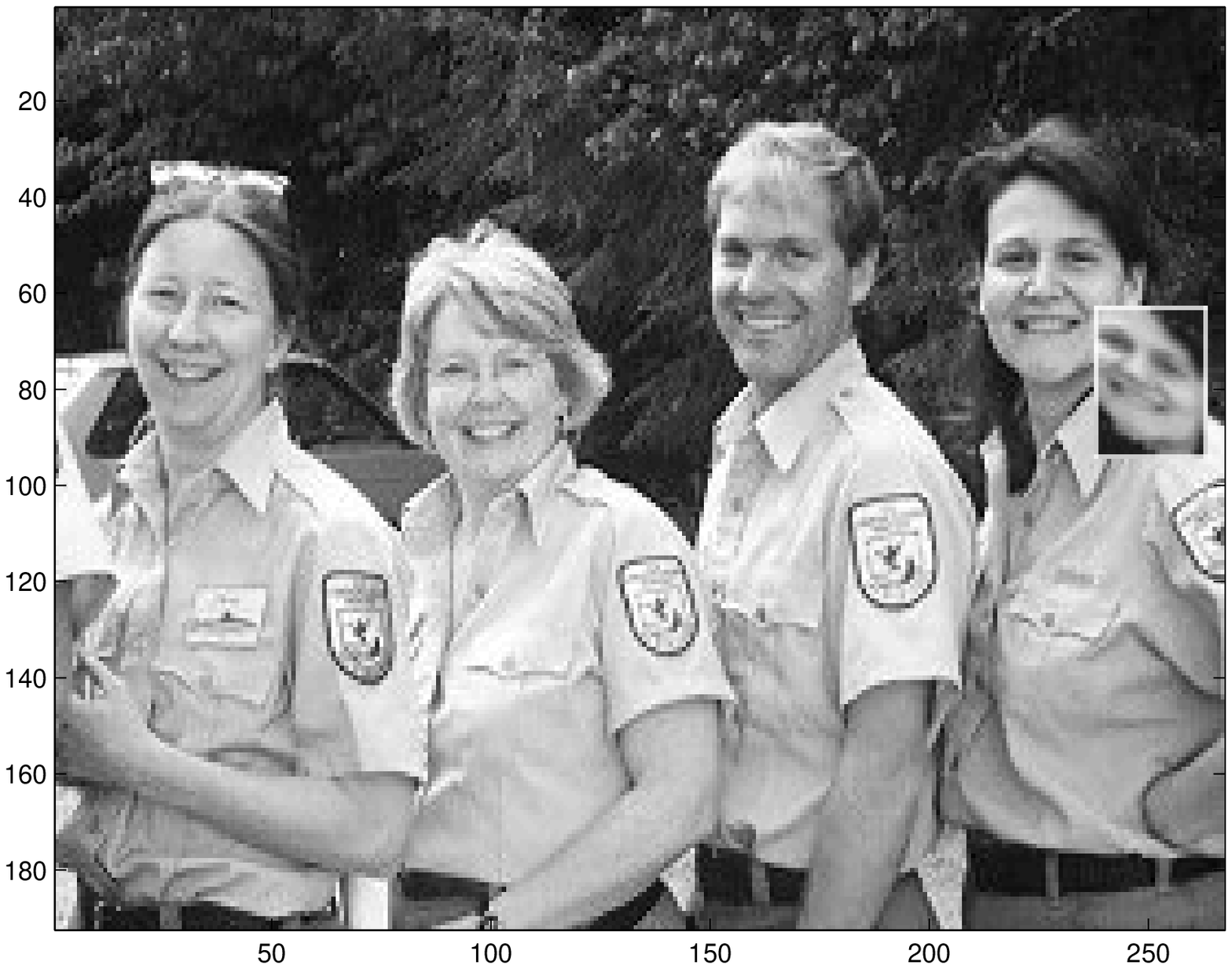,height=4cm}
  \hspace{10pt}
  &
  \psfig{figure=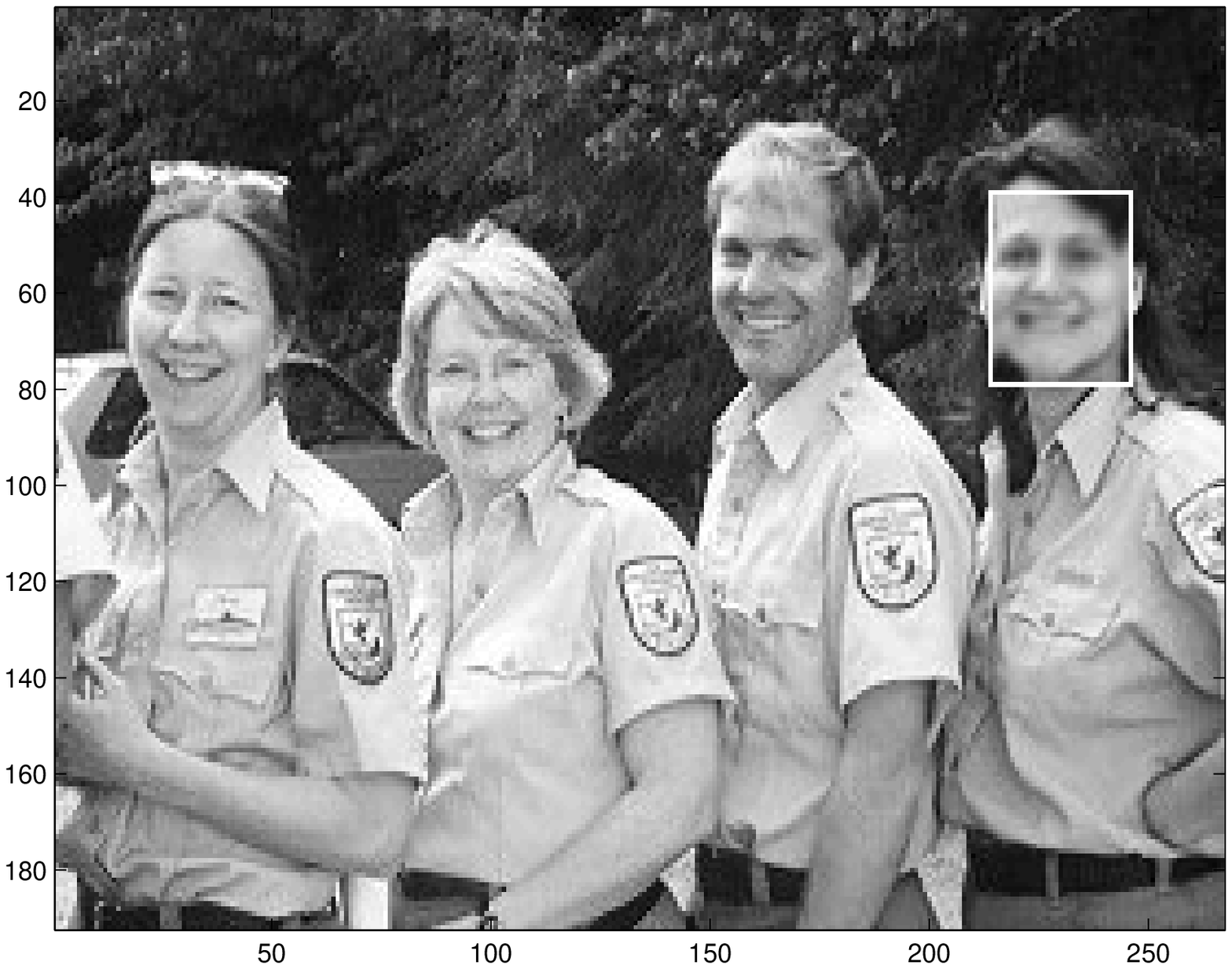,height=4cm}
\end{tabular}}
\caption{A randomly transformed template is placed on the image ({\it left}), and by means of numerical optimisation we find the parameters for which the log-posterior has a minimum value. The results can be seen on the ({\it right})}
\label{fig:Results2}
\end{figure}

We have experimented with greyscale images of faces, such as those shown in Fig. \ref{fig:Results1} and \ref{fig:Results2} . First, we present the effects of an appropriately chosen prior on the error function. In this example, we have isolated the scale space by choosing a rectangular template (the female face on the bottom right of the picture) and varying the scale parameters $s_{x}$,$s_{y}$ while keeping all other parameters constant. The resulting sum of square differences error (normalised to a value of 1) can be seen on the top-right of Fig. \ref{fig:Results1}. Note, that the desired solution is at $s_{x}$=$s_{y}$=$1$ and trivial solutions are located at values of either of the parameters $s$ close to zero. If we now choose a Wald prior, with a peak at $(1,1)$ (bottom-left), and calculate the inverse log-probability, we get the surface on the bottom-right. The trivial solutions have now become maxima, and the global minimum is at the desired solution $(1,1)$. Compared to the original function, the log-posterior surface is convex with a very large basin of attraction.
We also show some optimisation results, where a template is taken from the image (Fig. \ref{fig:Results2}), and is randomly affine transformed and locally deformed. We then use numerical optimisation to match the deformed template to the original image and see if we can find the correct parameters of the transformation. The template is placed on the image (Fig. \ref{fig:Results2}, left) and an exhaustive search is used on the translation parameters $d_{x}$,$d_{y}$, in order to find a good starting location for the optimisation algorithm. Using, for simplicity, a variation of the Simplex algorithm \cite{NelMea1965}, we minimise the parameters $\xi$ and obtain the resulting template which is superimposed on the right image of Fig. \ref{fig:Results2}. Visually, the results are quite pleasing, with the affine parameters being correctly identified within an appropriate error deviation (see Table \ref{table:Results_data}). 
 \section{Conclusions and Future Work}
We have presented a robust treatment of the view-oriented object recognition problem for intensity images under a Bayesian formulation. We have introduced prior distributions to bias appropriately a template which is deforming under affine transformation and a sinusoidal geometric deformation. Also, we have addressed the problem of different distributions of the foreground and background by using the robust smooth Huber metric. Some preliminary results obtained with our methods were presented.\begin{table}
\begin{center}
\caption{Comparison between actual and estimated values of the transformation parameters from Fig. \ref{fig:Results2}}
\label{table:Results_data}
\begin{tabular}{llll}
\hline\noalign{\smallskip}
Transformation & Actual & Estimated & Absolute deviation\\
\noalign{\smallskip}
\hline
\noalign{\smallskip}
Rotation $(\vartheta)$  & $30.47^{o}$ & $29.7046^{o}$ & $0.7654^{o}$ \\
Translation $(d_{x},d_{y})$ &$211,37$ & $213,38$ & $2,1$ \\
Scale $(s_{x},s_{y})$ &$1.3077,1.1923$  &$1.3125,1.2719$ & $0.0048,0.0796$\\
Shear $(\varphi)$& $27^{o}$ & $24.6776^{o}$ & $2.3224^{o}$\\
Sinusoid $(\alpha,k)$& $1.96,0.0327$ & $0.0032,0.0069$ &$1.9568,0.0258$ \\
\hline
\end{tabular}
\end{center}
\end{table}

There are many issues that we would like to examine in future work. In particular, we have only discussed grey-level imagery. Extension to colour imagery is needed. In addition, we would like to experiment with other metrics, more closely related to what is known about the statistics of images of natural and man-made scenes \cite{HuaMum1999}. We would also like to experiment with explicit modeling of the foreground and background distributions from training samples, using a statistical mixture model. Finally, in this early stage of our work, we have not discriminated between {\it intrinsic} variations of the template, that is variations of the shape of the object that depend only on the properties of the object and {\it extrinsic} variation which may depend on the viewpoint \cite{DiaBux2004}. We hope to introduce models for the {\it extrinsic}, viewpoint variations in the future.

\subsubsection{Acknowledgement} The primary author has been supported by a grant from the EPSRC.


\end{document}